\documentclass[conference,a4paper]{IEEEtran}

\makeatletter

\def\ps@IEEEtitlepagestyle{%
  \def\@oddfoot{\mycopyrightnotice}%
  \def\@evenfoot{}%
}
\def\mycopyrightnotice{%
  {\footnotesize 978-1-6654-6159-7/22/\$31.00 ©2022 IEEE\hfill}
  \gdef\mycopyrightnotice{}
}

\usepackage[table]{xcolor}
\usepackage[utf8]{inputenc}
\usepackage[T1]{fontenc}
\usepackage{graphicx}
\usepackage{amsmath}
\usepackage{hyperref}
\usepackage{tikz}
\usepackage{todonotes}
\usepackage{svg}      
\usepackage{floatrow} 
\usepackage{booktabs}

\usetikzlibrary{positioning}

\usepackage{blindtext}
\usepackage{eso-pic}
\IEEEoverridecommandlockouts
\usepackage{cite}
\usepackage{amsmath,amssymb,amsfonts}
\usepackage{algorithmic}
\usepackage{graphicx}
\usepackage{textcomp}
\def\BibTeX{{\rm B\kern-.05em{\sc i\kern-.025em b}\kern-.08em
    T\kern-.1667em\lower.7ex\hbox{E}\kern-.125emX}}
    
\usepackage{eso-pic}
\newcommand\AtPageUpperMyright[1]{\AtPageUpperLeft{%
 \put(\LenToUnit{0.17\paperwidth},\LenToUnit{-2cm}){%
     \parbox{0.9\textwidth}{\raggedleft\fontsize{8}{11}\selectfont #1}}%
 }}%
\newcommand{\conf}[1]{%
\AddToShipoutPictureBG*{%
\AtPageUpperMyright{#1}
}
}    

\begin{document}
\title{\vspace*{1cm}Hands-on detection for steering wheels with neural networks}

\author{\IEEEauthorblockN{Michael Hollmer}
\IEEEauthorblockA{%
	Faculty of Computer Science\\
	Deggendorf Institute of Technology\\
	Dieter-Görlitz-Platz 1\\
	94469 Deggendorf\\
	E-Mail: 
	\href{mailto:michael.hollmer@stud.th-deg.de}{michael.hollmer@stud.th-deg.de}
	}
	\and
\IEEEauthorblockN{Andreas Fischer}
\IEEEauthorblockA{%
	Faculty of Computer Science\\
	Deggendorf Institute of Technology\\
	Dieter-Görlitz-Platz 1\\
	94469 Deggendorf\\
	E-Mail: 
	\href{mailto:andreas.fischer@th-deg.de}{andreas.fischer@th-deg.de} (Corresponding author)
	}
}

\maketitle
\conf{\textit{  Proc. of the Interdisciplinary Conference on Mechanics, Computers and Electrics (ICMECE 2022)  \\ 
6-7 October 2022, Barcelona, Spain}}
\begin{abstract}
	In this paper the concept of a machine learning based hands-on detection algorithm is proposed. The hand detection is implemented on the hardware side using a capacitive method. A sensor mat in the steering wheel detects a change in capacity as soon as the driver's hands come closer. The evaluation and final decision about hands-on or hands-off situations is done using machine learning. In order to find a suitable machine learning model, different models are implemented and evaluated. Based on accuracy, memory consumption and computational effort the most promising one is selected and ported on a micro controller. The entire system is then evaluated in terms of reliability and response time.
\end{abstract}

\section{Introduction}
The development of advanced driver assistance systems is an essential goal for
car manufacturers. As can be seen from a survey, driver assistance systems are
by now an important purchase criterion for over 60\% of potential buyers
\cite{statista}. In addition, a unique selling point over the competition and
thus a competitive advantage can be gained through the further automation of
vehicles. An example is the system from Mercedes-Benz, which was the first to
receive approval for autonomous driving at level 3 in December 2021. 

Autonomous driving at level 3 enables the driver to divert his attention from
what is happening on the road in certain situations. The vehicle takes over the
lateral and longitudinal guidance and independently recognizes errors or
departure from system limits. In such a case, the system would
prompt the driver to take back control of the vehicle. This transfer of
vehicle control is a crucial challenge. An autonomous system must be able to
recognize whether the driver is ready to take over control of the vehicle
again. To ensure this, some form of driver monitoring is required. One way of
detecting the driver's condition is a hands-on detection (HOD). This is a
system that detects whether the driver's hands are on the steering wheel and
therefore control over the vehicle can safely be transferred.

A HOD can be implemented inexpensively by measuring steering angle and torque acting on the
steering wheel. The necessary sensors are required for the servo-assistance, anyway.
However, there is the disadvantage that false hands-off
messages often occur in situations where the driver does not
exert any significant force for lateral guidance. In such a case, the driver
would be asked to put his hands back on the steering wheel, even though he has
not let go of the steering wheel.

A better HOD variant, also used in this paper, uses a capacitance sensor. This
allows to detect the driver's contact with the steering wheel, without relying
on any exerted force to the steering wheel. However, the evaluation of
capacitance values is more complex, since these are dependent on the
driver and his environment. 

In this paper a machine learning algorithm is implemented, which is able to
distinguish between a hands-on and a hands-off situation based on the
capacitance values. The AI model is then ported to a micro controller and the
reliability and response time of the HOD is evaluated. A maximum response time of 200ms
is assumed to be appropriate for timely HOD. This paper aims to answer the
question: Can neural networks increase reliability of HOD within a response
time of 200ms?



\section{Background}
Two techniques are combined in this paper to realize HOD: Capacity measurement
and machine learning.

\subsection{Capacity measurement}

One option to realize HOD is detection of a contact between the driver and the
steering wheel by measuring the change in capacitance. There are different
methods to measure the capacitance of the steering wheel. In this paper, a
frequency-based measurement method is used. 
Touching the steering wheel is detected by a change in capacitance in a sensor
element, with the capacitance being calculated indirectly from the measured
frequency. The sensor element represents a measuring capacitor which forms a
resonant circuit together with another capacitor and a coil. The frequency of
the resonant circuit can be calculated using
equation~\ref{eq:ResonantFrequency}, which describes an ideal resonant circuit.
\begin{equation}
\label{eq:ResonantFrequency}
f_0 = \frac{1}{2 \pi \sqrt{L(C_k + C_s)}}
\end{equation}

The equation depends on the capacitance of the capacitor $C_k$, the capacitance
of the sensor element $C_s$ and the inductance of the coil $L$. As long as the
steering wheel is in an untouched state, the resonant circuit oscillates with
its maximum frequency $f_0$. If the driver puts his hands on the steering
wheel, the capacity of the sensor element is increased, leading to a
reduction in the frequency of the resonant circuit. 

The sensor element is a capacitive mat that is wrapped around the core of the
steering wheel and represents the active part of the measuring capacitor. Since
there is no opposite side, a stray electric field forms between the active
capacitor side and the environment. An approaching object causes a
change of the capacitance value of the sensor element.

To illustrate, the measuring capacitor can be seen as a plate capacitor, which can be described by the equation
$C = \epsilon_0 \cdot \epsilon_r \cdot \frac{A}{d}$.
Here, both electrically conductive and non-conductive objects
cause a change in capacitance for different reasons. A nearby conductive object
causes the distance $d$ between the active capacitor side and its surroundings
to decrease, increasing the capacitance. On the other hand, non-conductive
objects lead to an increase in capacity via a change in relative permittivity
$\epsilon_r$.

\subsection{Machine learning approaches}
To classify the capacitance values four different machine learning models are trained. 
In the following a brief overview of the different approaches is given.




\subsubsection{Time Delay Neural Network}
One machine learning approach is the Time Delay Neural Network (TDNN). The TDNN is structured as a standard multiperceptron with a delay buffer connected in front. New values in a time series are buffered until a certain amount is reached. Subsequently, these buffered values are passed in a final input into the multiperceptron, which then carries out the classification \cite{alpaydin_maschinelles_2019}. 

\subsubsection{Long Short Term Memory}
As a second approach to classify the capacitance values a Long Short Term Memory (LSTM) net, a variant of a recurrent neural network is used. In difference to feed-forward networks like the TDNN, the neurons of the LSTM can have connections to neurons in the previous layer, to the same layer or to themselves, in addition to the standard forward-pointing connections. The feedback loops implement a memory, which allows the network to remember previous events \cite{alpaydin_maschinelles_2019}. This is an advantage in time-dependent series of measurements, since each measured value is dependent on its predecessor in a certain way. In contrast to TDNN, which assumes independent measured values, recurrent neural networks can use this memory to take account of the temporal dependency \cite{hosseini_deep_2020}. 

\subsubsection{Random Forest}
The last approach is the random forest which combines the prediction results of multiple decision trees using the bootstrap aggregating (bagging) method. The idea behind bagging is to train several decision trees with a subset of the training data. The subsets are created by randomly selecting samples from the entire training data. This process is also called bootstrapping \cite{kubat_introduction_2021}. 
The result are multiple decision trees that are structured differently and ideally even out in their classification errors. The output of the random forest is the class chosen by most of the decision trees.

\section{Related Work} Other work also used machine learning to develop HOD,
differing in sensors, algorithms, and response times.

Johansson and Linder~\cite{johansson_system_2021} used a camera system and the
torque acting on the steering wheel to implement HOD. For the camera, two CNN
approaches were compared in classifying the most recently acquired image.  For
evaluating the torque measurement a one-dimensional convolutional neural
network and an LSTM network were used. According to the authors, the evaluation
of the torque requires a few seconds to detect a hands-off situation and up to
two seconds to detect a hands-on. The camera approach reacted to a situation
change within 5.4 seconds. Both solutions are thus well above the response time
of 200ms we aim for in this work.

Hoang Ngan Le et al.~\cite{Le_2016_CVPR_Workshops} have also developed a
machine learning based HOD with a camera system. In their paper the image
evaluation is performed by a Region Based Convolutional Neural Network (RCNN)
which has been improved for the specific purpose. 
The improved RCNN achieved 0.09 frames per second, which roughly corresponds to
the evaluation of one frame every eleven seconds.  As such, the time required
for detection is also well above the 200ms limit.

A solution not based on machine learning was published as a patent by
Volkswagen AG. This connects two possible approaches for HOD. On the one hand,
the values of the steering angle or torque sensor are used and on the other
hand, the capacitance values of the steering wheel are considered to
distinguish between a hands-on and hands-off situation. The idea behind the
combined approach is to use the torque sensor to detect hands-on situations
with high confidence. During these situations, the corresponding capacitance
values are recorded. With the data a function is set up with which it is
possible to quickly decide for each new capacitance value whether it
corresponds to a hands-on or hands-off
situation~\cite{musial_marek_dr_hands--erkennung_nodate}.

Another non machine learning option for evaluating capacitance sensors was
published by Analog Devices~\cite{ad7143_2007} and relies on dynamic threshold
values. An algorithm continuously monitors the values of the capacitance sensor
and measures the ambient level if no touch is detected. In addition, the
average maximum sensor value is measured with each touch. The threshold from
which a capacitance increase is counted as a touch is a certain percentage of
the average measured maximum sensor value. 

These approaches bear a potential problem: If the driver only touches the
steering wheel very lightly, the measured average maximum sensor value
decreases. The dynamic threshold adapts to the small capacitance values. Thus,
at some point a slight increase in capacitance values is erroneously recognized
as a touch. If the driver brings both hands close to the steering wheel without
touching it, this could trigger a similar increase in capacity as previous
two-finger touch. The HOD would then recognize a hands-on situation even though the
driver is not touching the steering wheel.

\section{Experimental description}
The implementation of the different machine learning models is divided into several steps. First, training data is recorded, classified and processed, which is then used to train the machine learning models. Based on the model results, the most promising model is selected and transferred to a micro controller. Finally, the system is evaluated in terms of reliability and response time.

\subsection{Generating training data}
\begin{figure}
    \centering
    \includegraphics[width=4cm]{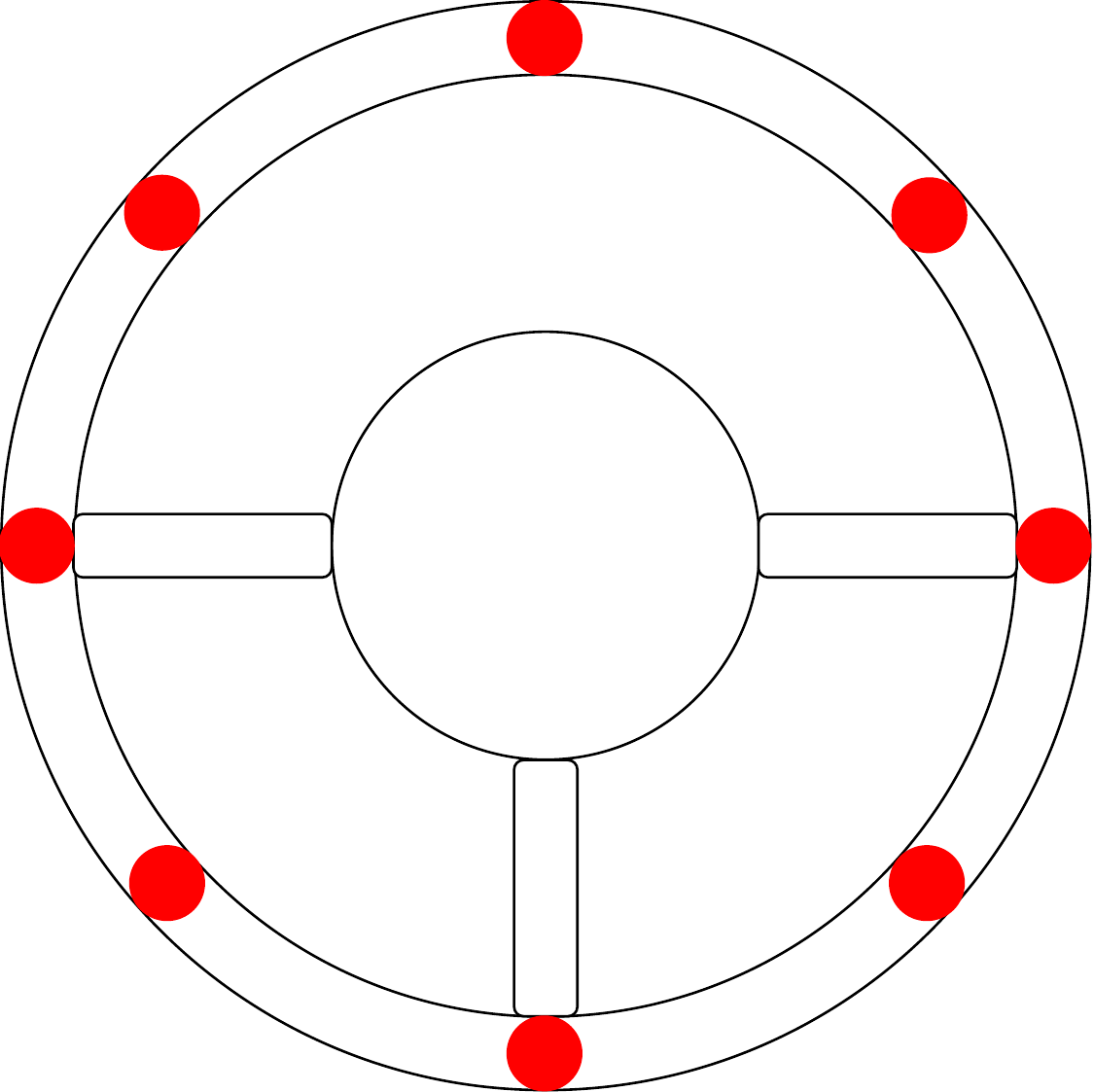}
    \caption{Contact points on the steering wheel during the training phase}
    \label{fig:ContactPoints}
\end{figure}

For generating the training data, the steering wheel is alternately touched and released at defined points for five seconds. This process is repeated 30 minutes each for a two-finger, four-finger and two-hand touch. Figure \ref{fig:ContactPoints} shows the points of contact. It should be noted that the points in the figure do not only refer to the front of the steering wheel. Alternately also the outside, the back and the inside were touched. Regarding the sampling rate a new capacitance value was recorded every 2 ms.

\subsection{Preprocessing data for learning}
After recording the training data two preprocessing steps were implemented. In the
first step, every sample was assigned a ``hands-on'' or
``hands-off' label. This was automated by following the change in capacitance
when touching or releasing the steering wheel. The corresponding edge was used to separate
and label all samples. Once the difference between two measured capacity values is above noise
level, it is interpreted as an edge. The required change in capacitance to trigger an edge
was set separately and fine tuned for each of the three data sets.
A rising edge triggers the ``hands-on'' label, while a falling edge triggers
the ``hands-off'' label.

In the second preprocessing step, every sample in the dataset was normalized in
its length. This was done because the machine learning model should learn to
classify a hands-on or hands-off situation based on capacitance values of just
a few hundred milliseconds to speed up the reaction time of the HOD. Therefore a
window with a fixed length of 100 values is placed over every sample. The
values in the window form the input for the machine learning models. In each
step, this window is moved one value, dropping an old value and adding a new
value. Thus, all models have a fixed length input of 100 capacitance values,
corresponding to 200 ms of recorded time. 


\subsection{Preparation of gradient data}

\begin{figure}
    \centering
    \includegraphics[width=7cm]{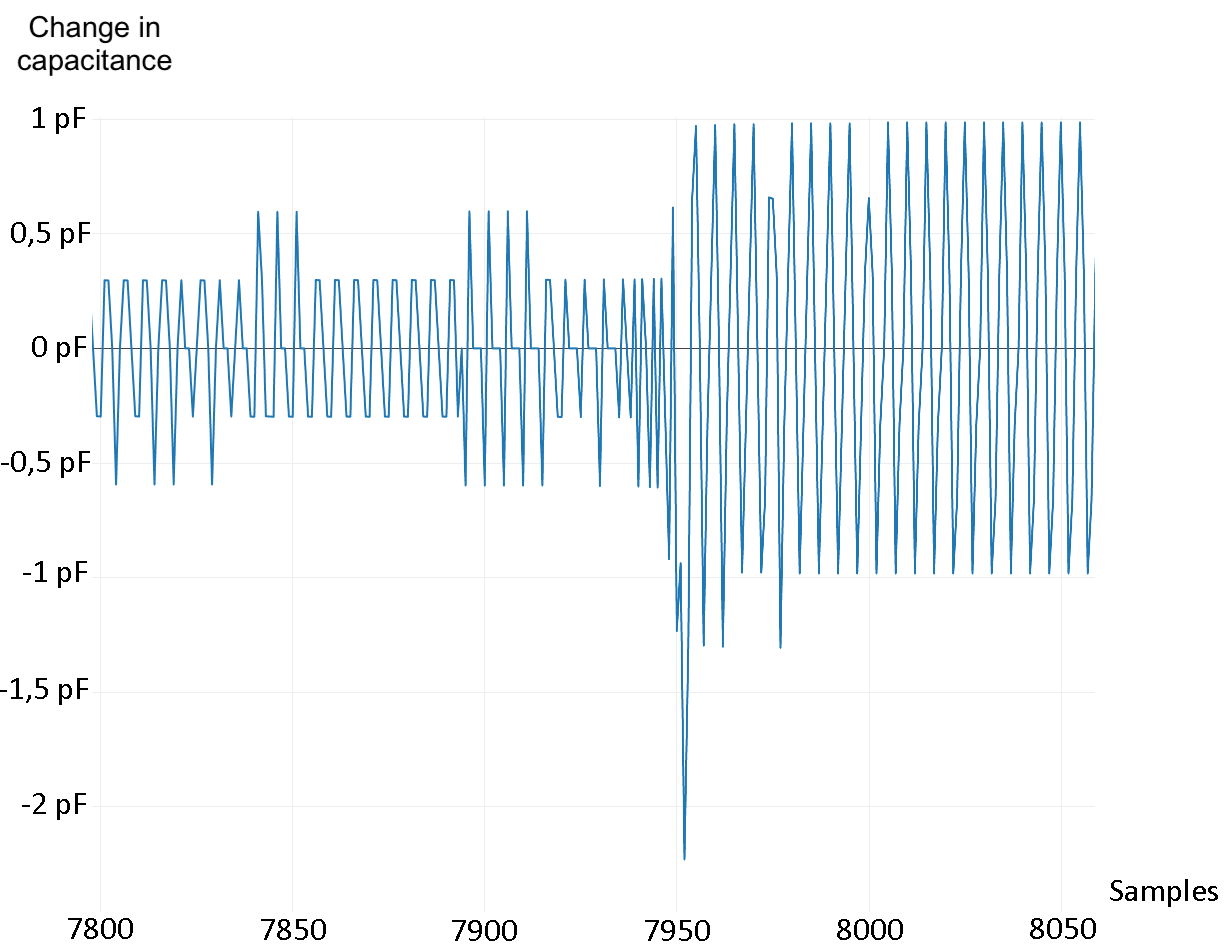}
    \caption{Change in capacity when the hand is approached}
    \label{fig:GradiantHandsOn}
\end{figure}

In order for the machine learning models to deliver optimal results,
capacitance values have to be normalized. For this, it is necessary to
obtain minimum and maximum capacitance value during execution. In
the training phase this is not a problem because all data is known a-priori. In
a real world application this is not the case, which means that minimum and
maximum values have to be determined dynamically. An estimate of the minimum value can
be obtained by measuring the ambient level when the steering wheel is
untouched. The maximum value, however, is a greater challenge. It would
require the driver to place both of his hands on the steering wheel, which the
system can never be sure is the case. Additionally, estimating the maximum value
from the minimum value is not possible, as the change in capacitance caused by
a driver heavily depends on his body weight. To eliminate this issue,
the absolute capacitance values were converted into gradient
values, focussing on change in capacity over time instead. This makes it easier
to normalize the values, since only the maximum capacitive rate of change need
to be known. Figure~\ref{fig:GradiantHandsOn} shows gradient values when
the steering wheel is touched with one hand.

\section{Evaluation}
For the evaluation all machine learning approaches are trained with the created datasets. The most promising model is then selected and ported to the STM32F769 micro controller where the final reliability and reaction time testing is done.
\subsection{Training}
In order to decide which machine learning model is best suited for the classification task, all models are trained with five different combinations of parameters. The resulting machine learning models are examined based on memory consumption, execution time and reliability. 

\subsubsection{Without gradient data}
\begin{table*}
	\centering
    \begin{tabular}[]{cccccccc}
        \toprule
        Model & Hidden Neurons & Accuracy & Precision & Recall & F0,5-Score & Memory & Exec.~time \\
        \midrule
        TDNN & 1 &  99,28\% & 98,78\% & 99,65\% & 98,95\% & 25kB & 12µs \\
        TDNN & 5 &  99,62\% & 99,53\% & 99,64\% & 99,56\% & 31kB & 21µs \\
        TDNN & 10 & 99,62\% & 99,53\% & 99,64\% & 99,56\% & 37kB & 34µs \\
        TDNN & 20 & 99,62\% & 99,54\% & 99,63\% & 99,56\% & 49kB & 59µs \\
        TDNN & 50 & 99,63\% & 99,56\% & 99,62\% & 99,57\% & 84kB & 150µs \\
        \midrule
        LSTM & 1 &  99,07\% & 98,35\% & 99,64\% & 98,61\% & 27kB & 0,6ms \\
        LSTM & 5 &  99,29\% & 98,87\% & 99,59\% & 99,01\% & 27kB & 3ms \\
        LSTM & 10 & 99,57\% & 99,48\% & 99,58\% & 99,50\% & 31kB & 7ms \\
        LSTM & 20 & 99,60\% & 99,48\% & 99,64\% & 99,51\% & 48kB & 15ms \\
        LSTM & 50 & 99,60\% & 99,49\% & 99,63\% & 99,52\% & 150kB & 59ms \\
        \midrule
        Model & Estimators & Accuracy & Precision & Recall & F0,5-Score & Memory & Features\\
        \midrule
        RF & 1   & 99,39\% & 99,45\% & 99,43\% & 99,44\% & 146kB & 10\\
        RF & 5   & 99,62\% & 99,66\% & 99,64\% & 99,65\% & 723kB & 10\\
        RF & 10  & 99,63\% & 99,68\% & 99,64\% & 99,67\% & 1.459kB & 10\\
        RF & 100 & 99,64\% & 99,70\% & 99,63\% & 99,69\% & 14.444kB & 10\\
        RF & 5   & 99,62\% & 99,67\% & 99,64\% & 99,66\% & 668kB & 15\\
        \bottomrule
    \end{tabular}
	\caption{Results of the TDNN, LSTM and the Random Forest (RF) after training with the recorded capacity data}
	\label{tab:PerformanceComparision}
\end{table*}

First, the models are trained with absolute capacitance values.
All models achieve a very high
level of accuracy as well as precision and recall, with differences visible mainly
in memory usage and execution time
(cf. Tab.~\ref{tab:PerformanceComparision}).
With no major difference in accuracy,
the random forest requires far more memory than other models, which
is particularly disadvantageous for embedded systems. Therefore,
measuring the execution time was neglected.

Looking at the neural networks, the biggest difference is the execution time.
While the TDNN only need a few microseconds for a forward pass, the LSTM
networks need several milliseconds due to their more complex structure. This is
relevant, because data is sampled with a rate of 2ms in the experiments.
Thus, when used on the micro controller, LSTM networks with
more than one hidden neuron lose data because the measurement is faster than
the processing.

\subsubsection{With gradient data}
Looking at the models trained with the gradient data, the previously observed
disadvantages regarding the memory consumption of the random forest remain
(cf.~Tab.~\ref{tab:PerformanceComparisionGradiant}). The
processing time of the LSTM is still inferior to that of the TDNN. However, the
LSTM with a hidden neuron performs slightly better in accuracy, precision and
recall compared to the TDNN with 50 hidden neurons and occupies with 27 kB of
memory almost just a third of the memory. The TDNN on the other hand, offers a
significantly shorter execution time. The delay between input and output for
the TDNN with 50 hidden neurons is only 150 $\mu$s compared to the 0.6 ms of
the smallest LSTM. Training the TDNN takes 1:08 minutes which is only a
fraction of the training time for the LSTM, which takes 20:48 minutes. 
For this reason the TDNN with 50 hidden neurons was selected and ported to the micro controller. 
\begin{table*}
	\centering
    \begin{tabular}[]{cccccccc}
        \toprule
        Model & Hidden Neurons & Accuracy & Precision & Recall & F0,5-Score & Memory & Exec.~time \\
        \midrule
        TDNN & 1  & 93,19\% & 98,81\% & 92,69\% & 97,52\%& 25kB & 12µs \\
        TDNN & 5  & 93,18\% & 99,50\% & 96,14\% & 98,81\%& 31kB & 21µs \\
        TDNN & 10 & 93,68\% & 99,87\% & 97,10\% & 99,30\%& 37kB & 34µs \\
        TDNN & 20 & 95,64\% & 99,84\% & 97,74\% & 99,41\%& 49kB & 59µs \\
        TDNN & 50 & 99,08\% & 99,64\% & 98,63\% & 99,44\%& 84kB & 150µs \\
        \midrule
        LSTM & 1  & 99,86\% & 99,79\% & 100\%   & 99,83\%& 27kB & 0,6ms \\
        LSTM & 5  & 99,84\% & 99,86\% & 99,80\% & 99,85\%& 27kB & 3ms \\
        LSTM & 10 & 99,66\% & 99,80\% & 99,66\% & 99,78\%& 31kB & 7ms \\
        LSTM & 20 & 98,23\% & 99,89\% & 99,72\% & 99,86\%& 48kB & 15ms \\
        LSTM & 50 & 99,82\% & 99,88\% & 99,63\% & 99,83\%& 150kB & 59ms \\
        \midrule
        Modell & Estimators & Accuracy & Precision & Recall & F0,5-Score & Memory & Features \\
        \midrule
        RF & 1   & 95,51\% & 94,88\% & 94,59\% & 94,82\%& 165kB & 10\\
        RF & 5   & 98,96\% & 98,41\% & 99,13\% & 98,55\%& 872kB & 10\\
        RF & 5   & 99,00\% & 98,50\% & 99,34\% & 98,67\%& 760kB & 15\\
        RF & 10  & 99,48\% & 99,29\% & 99,57\% & 99,35\%& 1.733kB & 10\\
        RF & 100 & 99,79\% & 99,61\% & 99,89\% & 99,67\%& 17.240kB & 10\\
        \bottomrule
    \end{tabular}
	\caption{Results of the TDNN, LSTM and the Random Forest (RF) after training with the calculated gradient data}
	\label{tab:PerformanceComparisionGradiant}
\end{table*}

\subsection{Practical reliability test}


To test if the system recognizes touches by the driver reliably, the steering
wheel was touched with two fingers, four fingers one hand and two
hands at the points shown in figure~\ref{fig:ContactPoints}. In the runs in which the
steering wheel was touched with two and four fingers, a distinction was made
between touching the front, back, inside and outside for each point. In each of
the four runs, all points were touched ten times to see whether touches were
only recognized sporadically in some places.


In these experiments, the recognition of two fingers proved to be the most
difficult. Especially on the inside, where there is a seam, the distance to the
sensor mat is particularly large. This decreases sensitivity and a touch
triggers only a small increase in capacitance, resulting in no touch detection
at all for the two finger experiments and a maximum of 7 out of 10 correctly
identified events in the four finger experiment. 

Regarding the position, the 6 o'clock position proved to be difficult, both with
two and with four fingers. Somewhat less (but still noticeably) impacted
positions were the 3 o'clock and 9 o'clock positions. These three positions are located,
where the steering wheel spokes connect to the wheel---likely the root causes of the problem.

In the 10 and 2 positions typical for driving a car, all events were recognized
reliably irrespective of finger count, as long as any area apart from the
inside of the wheel was touched. Also, when the steering wheel was not just
touched, but gripped with one or two hands, the success rate rose to 100\%.

\subsection{Reaction time}

\begin{figure}
    \centering
    \includegraphics[width=7cm]{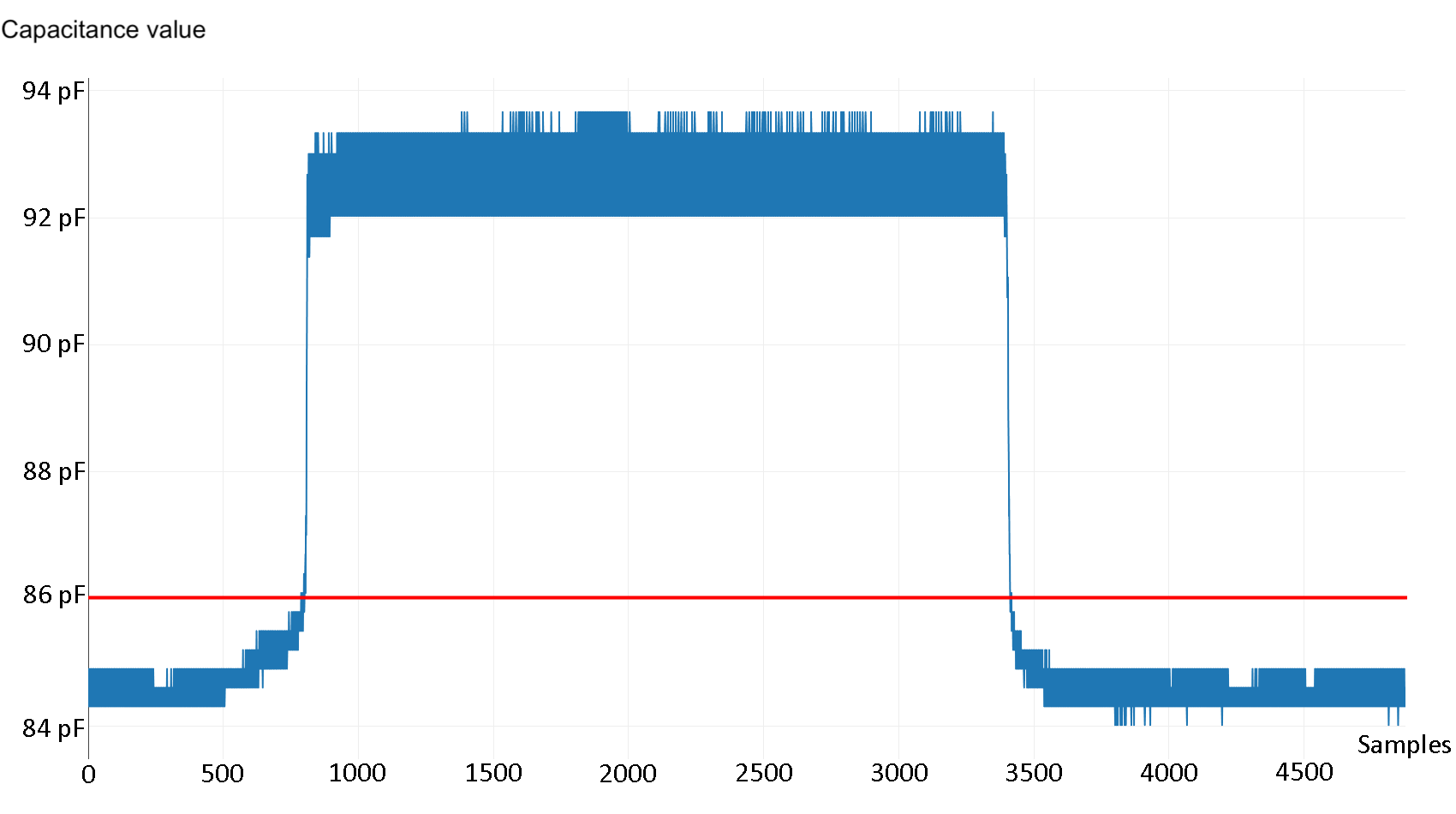}
    \caption{Touching the steering wheel with two fingers for the reaction time measurement}
    \floatfoot{The red line represents the threshold from which the time measurement is started and stopped}
    \label{fig:ReactionTime}
\end{figure}

Next, the reaction time of the system was tested with two fingers which
represents the hardest challenge as shown in the previous section.
Fig.~\ref{fig:ReactionTime} shows the capacitance values over time when the
steering wheel was touched with two fingers. The red line represents the
threshold from which the steering wheel was actually touched and released. The
increase in capacitance below the red line is caused by approaching the fingers
but not having made contact yet. Reaction time measurement is started, when the
capacitance values first exceed the threshold and stopped when the values drop
below it. In ten experimental runs, results ranged between 108ms and 294ms for
``hands-on'' events and a significantly faster reaction time of 30--60ms for
``hands-off'' events, if two fingers were used. With four fingers, reaction
times could be reduced to 74--94ms (hands-on) and 38--58ms (hands-off),
respectively.

\section{Conclusion}
The results show that it is possible to use a machine learning algorithm to evaluate capacitance values for HOD and achieve fast reaction times. By using the change in capacitance instead of the absolute values in the machine learning model, the problem of normalizing the input values was solved and the HOD worked without external calibration, independent of the driver and environment.

\bibliographystyle{IEEEtran}
\bibliography{bibliography}

\end{document}